\documentclass{article}

\usepackage[dblblindworkshop, final]{neurips_2026}
\workshoptitle{PhysWorldAI}

\usepackage[utf8]{inputenc} 
\usepackage[T1]{fontenc}    
\usepackage{hyperref}       
\usepackage{url}            
\usepackage{booktabs}       
\usepackage{amsfonts}       
\usepackage{nicefrac}       
\usepackage{microtype}      
\usepackage{xcolor}         
\usepackage{amsfonts}
\usepackage{amssymb}
\usepackage{graphicx}
\usepackage{amsmath}
\usepackage{booktabs}
\usepackage{array}
\usepackage{caption}
\usepackage{graphicx}
\usepackage{booktabs}
\usepackage{pifont}
\usepackage{multirow}
\usepackage{booktabs}
\usepackage{float}
\usepackage{xcolor}
\usepackage{colortbl}
\usepackage{xcolor}
\usepackage{xspace}
\usepackage{colortbl}
\usepackage{multirow}
\usepackage{soul}
\usepackage{multirow}
\usepackage{capt-of}
\usepackage[table]{xcolor}
\usepackage{colortbl}
\usepackage{booktabs} 
\usepackage{graphicx} 
\newcommand{\best}{}  
\title{Sparse-GS2Mesh: 3D Gaussian Splatting Guided by Novel Stereo Views and 2DGS for Sparse View Surface Reconstruction}

\author{%
  Younghyun Noh \\
  KT\\
  \texttt{younghyun.noh@kt.com} \\
  \And
  Minje Kim \\
  KAIST \\
  \texttt{minjekim@kaist.ac.kr} \\
  \And
  Tae-Kyun Kim \\
  KAIST \\
  \texttt{kimtaekyun@kaist.ac.kr} \\
}

\begin{document}
\maketitle

\begin{abstract}
Surface reconstruction under sparse-view settings remains challenging due to limited geometric cues. Volume rendering methods based on signed distance functions often produce over-smoothed surfaces, while 3D Gaussian Splatting (3DGS), though time-efficient, suffers from incomplete geometry due to the lack of reliable depth supervision and the limitation of being optimized only from given input views.
    In this paper, we present Sparse-GS2Mesh, a stereo-aware framework for surface reconstruction from sparse views.
    While 3DGS and stereo matching have been leveraged for surface reconstruction under dense view settings, we extend them to operate effectively under sparse view conditions by first initializing 3DGS using epipolar depth priors to mitigate the 3DGS overfitting problem, followed by our three key components: (I) adaptive baseline selection, (II) fine-tuning with a stereo matching network, and (III) 2D/3D co-regularized fine-tuning.
    Given a warmed-up 3DGS initialized with epipolar depth, the adaptive baseline selection automatically determines a baseline to synthesize for each sparse view. 
    We then fine-tune 3DGS by backpropagating depth-refining gradients from the stereo matching network, effectively specializing the 3DGS for stereo matching. 
    The 2D/3D co-regularization further helps obtain stable reconstruction, addressing weak geometric cues in close stereo views.
    Sparse-GS2Mesh achieves a 15\% improvement over state-of-the-art methods in little-overlap settings and comparable results in large-overlap settings.
    Project page: \url{https://yhnoh8623.github.io/projects/s-gs2mesh/}.
\end{abstract}

\begin{center}
\centering
\captionsetup{type=figure}
\includegraphics[scale=0.38]{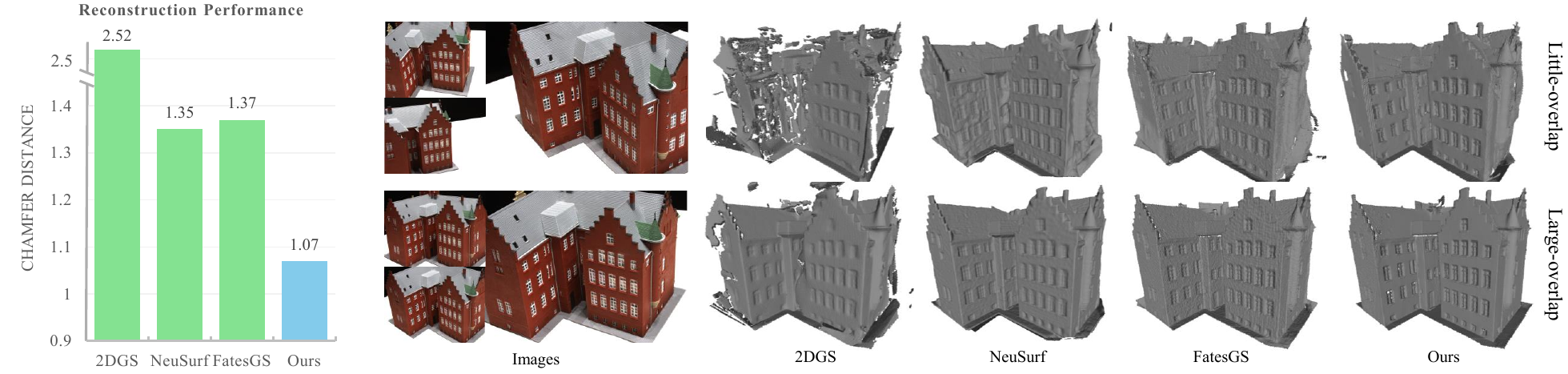}
\caption{
    Given sparse views as input, our method reconstructs high fidelity 3D geometry with novel stereo-view rendering and 2D/3D GS co-regularization.
}
\label{Teaser}
\end{center}

\section{Introduction}
\label{sec:intro}
3D surface reconstruction is a long-standing problem and plays a crucial role in various domains such as autonomous driving \cite{li2023bevdepth, zhuopeng2023read}, robotics \cite{qiyu2023graspnerf, zihan2022niceslam, moreau2021lens}, and virtual reality (VR) \cite{weicai2021superplane, matt2023objaverse}.
Despite recent progress, reconstructing accurate geometry from sparse inputs remains challenging yet practically important, since real-world applications often allow only a limited number of images. 

Previous surface reconstruction methods \cite{oechsle2021unisurf, tong2022voxurf, wang2023neus2} are built upon neural radiance fields (NeRF) \cite{mildenhall2020nerf}, representing a scene as continuous radiance/density fields and optimizing them via differentiable volume rendering.
Although several studies \cite{long2022sparseneus, ren2023volrecon, youngju2024uforecon, huang2024neusurf} have explored sparse-view reconstruction (i.e., only a few different view images as input) and achieved promising results, their major bottleneck lies in long training time, which ranges from several hours to days per scene.
Recent studies~\cite{yu2024gof, yu2024gsdf, zhang2024nsdgs} have extended 3D Gaussian Splatting (3DGS) \cite{kerbl20233dgs} to surface reconstruction.
2DGS \cite{huang20242dgs} directly reconstructs surfaces from 3DGS by flattening Gaussian elements and applying additional geometric regularization.
On the other hand, GS2Mesh \cite{yaniv2024gs2mesh} generates stereo-aligned image pairs through 3DGS rendering and extracts accurate depth using a pre-trained stereo model. 
In sparse-view settings, pre-trained multi-view stereo (MVS) networks \cite{xiong2023cl} have also been utilized \cite{jiang2025sparse2dgs} to initialize Gaussian elements and to enforce depth consistency during optimization.
Despite these developments, reconstructed surfaces in sparse-view settings still suffer from geometric inaccuracies due to extremely limited information.

To address the aforementioned challenges, we propose Sparse-GS2Mesh, a novel framework designed for surface reconstruction under sparse-view settings.
While the original GS2Mesh \cite{yaniv2024gs2mesh} was developed for dense-view settings, our key idea is that leveraging stereo-aligned renderings as additional geometric cues becomes more effective under sparse-view conditions, where geometric information is extremely limited.
However, in such sparse-view settings, 3DGS tends to overfit to training views, leading to a rapid degradation in novel-view rendering quality as the viewpoints move away from the training views.
Moreover, initializing 3DGS from sparse SfM points further intensifies overfitting, resulting in unstable Gaussians.

Therefore, our method initializes the 3DGS using epipolar depth priors following the strategy of NexusGS~\cite{yulong2025nexusgs}, to obtain more reliable stereo renderings in the early stage. 
Instead of using a fixed baseline as in prior works, we adaptively determine the baseline for each view using a multi-view feature score, avoiding the manual setup of baseline parameters.
Afterwards, we fine-tune the 3DGS using a pre-trained stereo matching network together with a 2D/3D co-regularization scheme.
The stereo matching network is leveraged to propagate gradients that directly improve its depth estimation quality, guiding the 3DGS toward stereo-consistent geometry.
Meanwhile, 2D/3D co-regularization provides geometrically stable optimization by complementing the strengths of 2D and 3D Gaussian representations.
It is particularly important in large-overlap settings where the synthesized stereo images become similar to the existing input images, making them redundant and thereby weakening the effect.
The fine-tuned 3DGS is specialized for stereo-view rendering, from which accurate depth maps are extracted using a pre-trained stereo network.
High-quality images rendered from the fine-tuned 3DGS offer additional geometric cues under sparse-view settings, improving our method to preserve more accurate geometry compared to prior works \cite{jiang2025sparse2dgs, han2025fatesgs}, which rely solely on information derived from the given sparse views. 
Experiments were conducted on the DTU dataset \cite{aanaes2016dtu}, and surface reconstruction qualities are evaluated using the Chamfer Distance metric.
We also perform qualitative experiments on the BlendedMVS \cite{yao2020blendedmvs} dataset to demonstrate the generalization of our method.
Our contributions are summarized as follows:
\begin{itemize}
    \item We propose Sparse-GS2Mesh, the pipeline specializing 3DGS for stereo matching under sparse inputs, with three novel components: (I) adaptive baseline selection, (II) fine-tuning with a stereo matching network, and (III) 2D/3D co-regularized fine-tuning.
\end{itemize}
\begin{itemize}
    \item We use synthesized stereo images from 3DGS as geometric cues and leverage epipolar depth priors in initialization to enhance reconstruction quality. 
\end{itemize}
\begin{itemize}
    \item Sparse-GS2Mesh achieves Chamfer Distance (mm) of 1.07 on the DTU dataset with three little-overlap images, outperforming the SOTA (1.27), and 0.95 under the large-overlap setting, which is comparable to the SOTA (0.92).
\end{itemize}

\section{Related Work}

\subsection{Dense View Surface Reconstruction}

Neural Radiance Fields (NeRF)~\cite{mildenhall2020nerf} represent 3D scenes implicitly for novel view synthesis via differentiable volume rendering. Subsequent studies extend this formulation to surface reconstruction; NeuS~\cite{peng2021neus} improves geometric fidelity via weighted volume rendering, while RegSDF~\cite{zhang2023regsdf} regularizes geometry using SfM point clouds. However, these radiance-field methods require heavy computation and degrade significantly under sparse-view conditions due to insufficient geometric cues.

3D Gaussian Splatting (3DGS)~\cite{kerbl20233dgs} models scenes using explicit Gaussian primitives optimized by photometric consistency. Recent works extend 3DGS to dense surface reconstruction: 2DGS~\cite{huang20242dgs} flattens Gaussians for surface alignment, GS2Mesh~\cite{yaniv2024gs2mesh} predicts depth from rendered stereo pairs using a pre-trained stereo model, and PGSR~\cite{chen2024pgsr} integrates unbiased depth rendering with cross-view regularization. Despite high quality under dense views, their dependence on dense observations limits performance in sparse scenarios, motivating our approach.

\subsection{Sparse-View 3D Reconstruction}
\paragraph{Novel View Synthesis.} 
Although both NeRF \cite{mildenhall2020nerf} and 3DGS \cite{kerbl20233dgs} achieve impressive performance in novel view synthesis under dense-view settings, they tend to overfit to training views when the number of input images is constrained, leading to a significant degradation in novel view rendering quality.
To alleviate this issue, DS-NeRF~\cite{deng2022dsnerf} leverages depth priors from SfM points to constrain scene reconstruction, and InfoNeRF~\cite{kim2022infonerf} reduces reconstruction uncertainty via entropy-based regularization on ray densities. 
Cor-GS~\cite{zhang2024corgs} extends 3DGS with a dual-field co-regularization scheme to enhance multi-view geometric consistency.
Unlike Cor-GS, our framework enforces co-regularization between 3DGS and 2DGS, leading to a more robust 3DGS representation, particularly well-suited for surface reconstruction.
NexusGS~\cite{yulong2025nexusgs} exploits geometric initialization by computing epipolar depth from optical flow estimation.
Our method follows the same initialization strategy as NexusGS, and further refines the geometry through fine-tuning using the stereo matching model and 2D/3D Co-Regularization.

\paragraph{Surface Reconstruction.} 
Beyond novel view synthesis, surface reconstruction also faces challenges under sparse-view settings due to limited geometric observations.
To address this issue, NeuSurf~\cite{huang2024neusurf} integrates on-surface priors into neural implicit representations, while SparseNeuS~\cite{long2022sparseneus} learns generalizable geometric priors from large-scale datasets using signed distance functions (SDF) and image features, though it still requires per-scene fine-tuning.
FatesGS~\cite{han2025fatesgs}, a 2DGS~\cite{huang20242dgs} based method, improves geometric alignment by enforcing both multi-view feature consistency and intra-view depth consistency.
Sparse2DGS~\cite{jiang2025sparse2dgs} further strengthens 2DGS by introducing a geometry-prioritized enhancement strategy that combines the advantages of MVS and 2DGS.
Despite these advances, their performance remains limited under extremely sparse observations (i.e., a little overlap setting), often yielding noisy or incomplete geometry. 
These methods rely solely on the information available from the given views without leveraging additional geometric cues.
This motivates our stereo-guided Sparse-GS2Mesh framework, which effectively leverages high-quality stereo cues to achieve robust geometry reconstruction under sparse-view conditions.

\subsection{Deep Learning Based Stereo Matching}

Stereo matching aims to estimate disparity from a pair of stereo images captured from closely spaced viewpoints. Recent deep-learning methods~\cite{chang2018pyramidstereo, shen2021cfnet, zhang2019ganet, wen2025foundationstereo}, particularly recurrent architectures~\cite{lipson2021raft, xu2023igev, zhao2023dlnr}, iteratively refine disparity predictions for high accuracy. Among them, DEFOM-Stereo~\cite{jiang2025defomstereo} integrates monocular depth estimation on the reference view with recurrent stereo refinement. To fully exploit the information from the ground-truth image alongside our synthesized 3DGS stereo rendering, we adopt DEFOM-Stereo to iteratively refine the monocular depth within its recurrent framework.

\section{Method}
We introduce Sparse-GS2Mesh, a method to reconstruct high-fidelity 3D geometry from sparse-view inputs. Our framework first goes through a 2D/3D GS~\cite{huang20242dgs, kerbl20233dgs} warm-up stage, and then adaptively selects stereo baselines without manual tuning to synthesize reliable stereo-aligned images. 
During the fine-tuning stage, the stereo matching network~\cite{jiang2025defomstereo} adapts 3DGS specifically for stereo matching, while the 2D/3D co-regularization improves their robustness by allowing two modules to compensate for each other's strengths.
Finally, we render high-quality stereo views from 3DGS and extract accurate depth maps using the pre-trained stereo matching network to obtain high-quality meshes. We introduce the 2D/3D GS warm-up stage in Sec~\ref{sec:warmup stage}, adaptive baseline selection in Sec~\ref{adaptive baseline determination}, and fine-tuning strategies in Sec~\ref{fine-tuning with stereo matching network} and Sec~\ref{fine-tuning with co-regularization}.

\begin{figure*}[!htbp]
\centering
\includegraphics[width=0.95\textwidth]{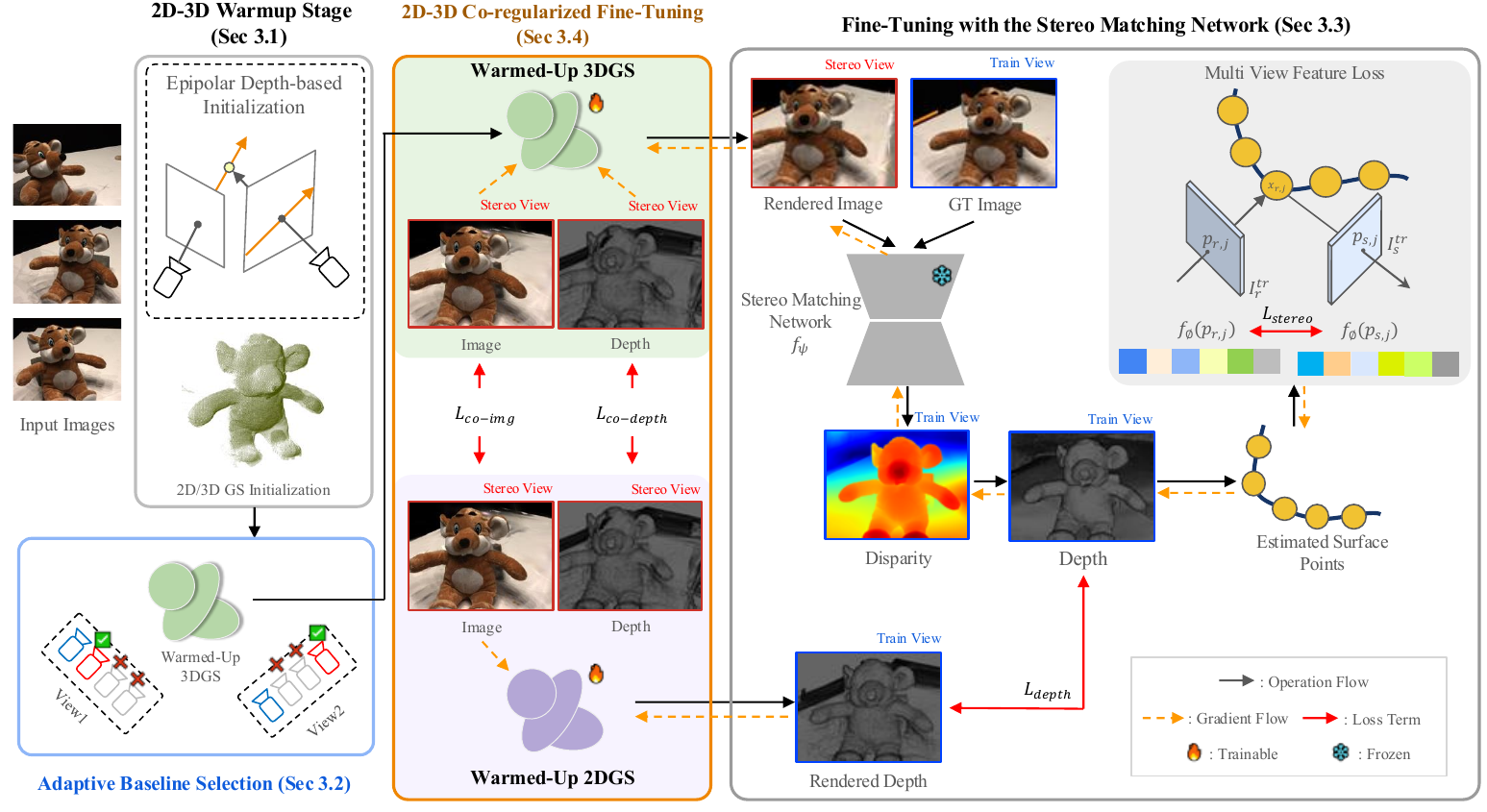}
\caption{\textbf{Overview of Sparse-GS2Mesh.} Stereo renderings from warmed-up 3DGS are refined via backpropagated gradients from the stereo matching network (Sec.~\ref{fine-tuning with stereo matching network}). Subsequently, 2D/3D co-regularization with stereo image and depth supervision stabilizes 3DGS geometry (Sec.~\ref{fine-tuning with co-regularization}), specializing it for stereo matching.}
\label{fig:fine-tuning_framework}
\vspace{-5pt}
\end{figure*}

\subsection{2D/3D Gaussian Splatting Warm-Up Stage}
\label{sec:warmup stage}
To enhance the performance of Gaussian Splatting under sparse-view conditions, we follow the NexusGS ~\cite{yulong2025nexusgs} initialization strategy, which leverages depth estimation based on epipolar geometry and optical flow.
This initialization alleviates the overfitting problem under sparse-view settings. 
During the warm-up training stage, 3DGS~\cite{kerbl20233dgs} is optimized solely with a photometric loss.
For the 2DGS~\cite{huang20242dgs} warm-up, we use the photometric loss together with the depth distortion loss and normal consistency loss to optimize surface geometry following the original 2DGS.

Warmed-up models are subsequently fine-tuned through the pre-trained stereo matching network (Sec \ref{fine-tuning with stereo matching network}) and 2D/3D co-regularization (Sec \ref{fine-tuning with co-regularization}). 
Hereafter, the term left view refers to the training view that contains the ground-truth image, and the right view refers to the synthesized stereo view.

\subsection{Adaptive Baseline Selection From Multi-View Feature Scores}
\label{adaptive baseline determination}
The stereo baseline, which represents the distance between two input cameras in a stereo pair, is a crucial hyperparameter that influences the performance of stereo matching. Previous work~\cite{yaniv2024gs2mesh} uses a fixed baseline (typically 7\% of the scene radius) and was empirically applied to all views. However, the optimal baseline varies across views depending on the geometry observed from each viewpoint, particularly in our setting where stereo views are synthesized from the 3DGS~\cite{kerbl20233dgs}.

To this end, we present an adaptive view-dependent baseline selection that can select baseline views without any manual settings. 
Given the left camera parameters ($R_L$,$T_L$), the corresponding right camera parameters ($R_R$, $T_R$) for a baseline $b$ are expressed as:
\begin{equation}
R_R = R_L, \quad T_R = T_L + R_L \cdot [b, 0, 0].
\label{eq:baseline}
\end{equation}
To determine the baseline, we define candidate baselines $\{B_i \mid 1 \le i \le N_B\}$ ranging from short to wide distances for each view, where $N_B$ denotes the number of candidate baselines. 
For each candidate $B_i$, we compute the corresponding right view (stereo view) camera parameters using Eq.\ref{eq:baseline} and render the stereo view image $I^{st}_i$ using warmed-up 3DGS from Sec. \ref{sec:warmup stage}.  
$I^{st}_i$ and the train view ground-truth image $I^{tr}_i$ are then fed into a pre-trained stereo matching network~\cite{jiang2025defomstereo} to estimate train view depth $\{D_i \mid 1 \le i \le N_B\}$.

The multi-view feature consistency is measured to evaluate the reliability of each candidate depth.
While previous works ~\cite{han2025fatesgs, huang2024neusurf} use multi-view feature consistency as a loss term, we reinterpret it as a score in this section.
Let $I^{tr}_{r}$ be the reference train view ground-truth image, and $I^{tr}_{s}$ be one of the source train view ground-truth images, both downscaled by a factor of $l$.
A pixel $p_{r,j}$ in $I^{tr}_{r}$ is unprojected to a 3D point using the candidate depth value $D_{i,j}$ and then reprojected onto the source-view pixel $p_{s,j}$ in $I^{tr}_{s}$ by
\begin{equation}
x_{r,j} = o_r + D_{i,j} \cdot d_j,
\label{eq:unprojection}
\end{equation}
\vspace{-15pt}
\begin{equation}
p_{s,j} = K P_s^{-1} x_{r,j},
\label{eq:projection}
\end{equation}
where K and $P_s$ are the intrinsic and extrinsic parameters of the source view.
$o_r$ is the reference view camera center, and $d_j$ is the normalized direction vector passing through $p_{r,j}$ originated from $o_r$.
We compute the cosine similarity between the feature embeddings of pixels $p_{r,j}$ and $p_{s,j}$, aggregating over all pixels, downscale factors, and source views to obtain the multi-view feature score $S_i$:
\begin{equation}
S_i = \sum_{s,j,l} 
\frac{
f_{\phi}(I^{tr}_{r,l},p_{r,j}) \cdot f_{\phi}(I^{tr}_{s,l},p_{s,j})
}{
\|f_{\phi}(I^{tr}_{r,l},p_{r,j})\| \, \|{f_{\phi}(I^{tr}_{s,l},p_{s,j})}\|
},
\label{eq:similarity}
\end{equation}
where $p_{r,j} \in M^{ref-src}_{occ}$, 
and $M^{ref-src}_{occ}$ is the occlusion mask between reference and source view derived from Eq.\ref{eq:unprojection} and Eq.\ref{eq:projection}.
$f_{\phi}(\cdot)$ denotes the feature extraction network ~\cite{zhang2020vismvsnet}, and $l$ increases in powers of two ($l \in \{1, 2, 2^2, \dots\}$).  
Finally, we determine the  baseline $B^*$ for each view as the one that yields the highest $S_i$:
\begin{equation}
B^* = \arg\max_{B_i} S_i.
\label{eq:argmax}
\end{equation}
This adaptive selection automatically determines the optimal baseline for each view, enabling accurate depth prediction.
The baselines obtained in this section are subsequently used in the fine-tuning stage (Sec \ref{fine-tuning with stereo matching network}, Sec \ref{fine-tuning with co-regularization}).  

\subsection{Fine-Tuning with the Stereo Matching Network}
\label{fine-tuning with stereo matching network}
Unlike conventional 3DGS~\cite{kerbl20233dgs} approaches that aim to render across arbitrary novel viewpoints, our framework focuses on the stereo views defined in Sec \ref{adaptive baseline determination}. 
The goal is not only to improve rendering quality, but also to produce effective inputs for the stereo matching network~\cite{jiang2025defomstereo}.
To this end, we fine-tune the 3DGS to align with the objective described above.
As illustrated in Fig.\ref{fig:fine-tuning_framework}, the stereo view rendering from the 3DGS is fed into the pre-trained stereo matching network, together with the train-view ground truth image.
The gradients that refine the network's depth are backpropagated through the stereo model into the 3DGS, enabling the 3DGS to adapt specifically for stereo matching rather than general rendering.

The loss applied to the estimated depth is designed to maximize the multi-view feature consistency defined in Sec \ref{adaptive baseline determination}. 
For the reference train view, we compute the corresponding stereo view using the previously obtained optimal baseline $B^*$ and Eq.\ref{eq:baseline}, and render $I^{st}_r$.
Using the reference train view ground-truth image $I^{tr}_r$ and the stereo view rendered image $I^{st}_r$, we compute the reference train view depth $D^{tr}_r$ using the pre-trained stereo matching network as
\begin{equation}
D^{tr}_r = f_{\psi}(I^{tr}_r, I^{st}_r).
\label{eq:stereo}
\end{equation}
where $f_{\psi}$ denotes the pre-trained stereo matching network~\cite{jiang2025defomstereo}.
Using $D^{tr}_r$ together with Eq.\ref{eq:unprojection} and Eq.\ref{eq:projection}, we obtain the corresponding pixel pairs ($p_{r,j}$, $p_{s, j}$) between the reference view ground-truth image and the source view ground-truth image in the same manner as described in Sec \ref{adaptive baseline determination}.
To encourage the model to predict more accurate depth, we employ a loss that maximizes the cosine similarity between the features of the corresponding pixel pairs.
The loss is formulated as follows:
\begin{equation}
L_{stereo} = \sum_{s,j,l} \Bigg(
1 -
\frac{
f_{\phi}(I^{tr}_{r,l},p_{r,j}) \cdot f_{\phi}(I^{tr}_{s,l},p_{s,j})
}{
\|f_{\phi}(I^{tr}_{r,l},p_{r,j})\| \, \|{f_{\phi}(I^{tr}_{s,l},p_{s,j})}\|
}
\Bigg)
,
\label{eq:multiview stereo feat loss}
\end{equation}
where $p_{r,j} \in M_{occ}=M^{ref-src}_{occ} \cap M^{tr-st}_{occ}$. 
Here, $M^{ref-src}_{occ}$ is from Sec \ref{adaptive baseline determination} and $M^{tr-st}_{occ}$ is the occlusion mask between training and stereo view derived from the stereo matching network.
The gradients produced by this loss are backpropagated through the stereo matching network into the 3DGS parameters. 
This enables the 3DGS to be fine-tuned in a manner specialized for stereo matching.

\subsection{2D/3D Co-Regularized Fine-Tuning}
\label{fine-tuning with co-regularization}
While fine-tuning with the stereo matching network already provides plausible reconstruction results, synthesized stereo images become similar to the other given input images when captured from adjacent viewpoints, especially in a large-overlap setting. 
This redundancy weakens the effectiveness of our motivation, as the geometric cues generated from the stereo image overlap with the existing information.
Under such conditions, the depth predicted by the stereo matching network might be less reliable than the depth rendered from the 2DGS ~\cite{huang20242dgs}.
To address this issue, we strengthen the 3DGS~\cite{kerbl20233dgs} through a co-regularization scheme with the 2DGS branch, which encourages mutual consistency between the two representations.
We demonstrate that this 2D/3D co-regularization improves reconstruction quality in both little- and large-overlap settings (see the experiments).

While 2DGS has a tendency to attach Gaussian elements tightly to the surface, causing less artifacts, it is less expressive in image rendering than 3DGS.
Through the proposed co-regularization, the two branches complement each other’s strengths, improving both geometric alignment and image fidelity.
As illustrated in Fig.\ref{fig:fine-tuning_framework}, we render the stereo view image $I^{st}_{3D}$ from the 3DGS and $I^{st}_{2D}$ from the 2DGS using the optimal baseline determined in Sec \ref{adaptive baseline determination}.
We then apply photometric loss between $I^{st}_{3D}$ and $I^{st}_{2D}$ to ensure consistent image appearance across representations.
The loss is formulated as follows:
\begin{equation}
\begin{split}
L_{\textrm{co-img}} 
= (1-\lambda)L_1(I^{st}_{3D}, I^{st}_{2D}) + \lambda L_{D\text{-}SSIM}(I^{st}_{3D}, I^{st}_{2D}).
\end{split}
\label{eq:co-img_loss} 
\end{equation}
Additionally, we render the stereo view depth $D^{st}_{3D}$ from the 3DGS and $D^{st}_{2D}$ from the 2DGS.
We apply $L_2$ loss between the depth renderings of the 3DGS and 2DGS, where the gradients are propagated only to the 3DGS branch.
This helps refine the 3DGS geometry even under large-overlap conditions where the stereo cues provide limited new information.
The loss is formulated as follows:
\begin{equation}
L_{\textrm{co-depth}} = L_2(D^{st}_{3D},D^{st}_{2D}).
\label{eq:co-depth_loss}
\end{equation}
Finally, we introduce another $L_2$ loss between the train view depth rendered from 2DGS and the depth estimated by the stereo matching network to ensure stable learning under various input conditions.
The loss is formulated as follows:
\vspace{-7pt}
\begin{equation}
L_{\textrm{depth}} = L_2(D^{tr}_{2 D},D^{tr}_r).
\label{eq:depth_loss}
\end{equation}

\subsection{Total Training Losses}
\label{Training Loss}
During the fine-tuning stage, in addition to the losses described above, the 3DGS branch is optimized with a photometric loss $L_{rgb}$, and the 2DGS branch is trained with the photometric loss, the 2D regularization loss composed of the normal consistency loss and depth distortion loss following the original 2DGS~\cite{huang20242dgs}. The final loss for the 3DGS and 2DGS branches are defined as:
\begin{equation}
\begin{aligned}
L_{3DGS} = L_{rgb} + L_{co} + \beta_{1}L_{stereo},\quad
L_{2DGS} = L_{rgb} + \beta_2L_{2d-reg} + L_{co} + \beta_3L_{depth},\\
where \quad L_{co} = \beta_4L_{co-img} + \beta_5L_{co-depth},
\label{eq:2dgs total loss}
\end{aligned}
\end{equation}
$\beta_1$, $\beta_2$, $\beta_3$, $\beta_4$, $\beta_5$ denote the weights of individual loss terms.
Although the full objective involves multiple terms, they are partitioned to the 2DGS and 3DGS branches, with the two terms shared across the both.

\section{Experiments}
\label{sec:blind}

In this section, we evaluate the performance of our method through both quantitative and qualitative experiments.
We describe the datasets used in our experiments and provide the implementation details in Sec \ref{dataset} and Sec \ref{implementation}, respectively.
Sec \ref{result} presents the surface reconstruction results in both quantitative and qualitative evaluations.
Finally, Sec \ref{ablation} provides comprehensive ablation studies to validate the contribution of each component in our framework.

\begin{table*}[t]
\centering
\renewcommand{\arraystretch}{1}
 \setlength{\tabcolsep}{5pt}
 \scalebox{0.65}{
\begin{tabular}{lcccccccccccccccc}
\toprule
Scan        & 24   & 37   & 40   & 55   & 63   & 65   & 69   & 83   & 97   & 105  & 106  & 110  & 114  & 118  & 122  & Mean \\ \midrule
Colmap~\cite{schonberger2016structure}      &2.88 &3.47 &1.74 &2.16 &2.63 &3.27 &2.78 &3.63 &3.24 &3.49 &2.46 &1.24 &1.59 &2.72 &1.87 &2.61 \\ \midrule
SparseNeus~\cite{long2022sparseneus}  & 4.81 & 5.56 & 5.81 & 2.68 & 3.30 & 3.88 & 2.39 & 2.91 & 3.08 & 2.33 & 2.64 & 3.12 & 1.74 & 3.55 & 2.31 & 3.34 \\
VolRecon~\cite{ren2023volrecon} & 3.05 & 4.45 & 3.36 & 3.09 & 2.78 & 3.68 & 3.01 & 2.87 & 3.07 & 2.55 & 3.07 & 2.77 & 1.59 & 3.44 & 2.51 & 3.03 \\
UFORecon~\cite{youngju2024uforecon}  & 1.52 & 2.58 & \underline{1.85} & 1.44 & 1.55 & 1.81 & 1.06 & 1.52 & \underline{0.96} & 1.40 & 1.19 & \underline{0.94} & 0.65 & 1.25 & 1.29 & 1.40 \\
NeuSurf~\cite{huang2024neusurf} & 1.35 & 3.25 & 2.50 & \best\textbf{0.80} & 1.21 & 2.35 & \underline{0.77} & \best\textbf{1.19} & 1.20 & 1.05 & \best\textbf{1.05} & 1.21 & \best\textbf{0.41} & \best\textbf{0.80} & 1.08 & 1.35 \\ \midrule
2DGS~\cite{huang20242dgs}  & 3.25 & 3.64 & 3.52 & 1.42 & 2.04 & 2.52 & 1.99 & 2.69 & 2.55 & 1.79 & 2.92 & 4.50 & 0.73 & 2.38 & 1.79 & 2.52 \\
GS2Mesh~\cite{yaniv2024gs2mesh}  & 2.60 & 3.15 & 3.16 & 1.24 & 1.64 & 2.37 & 1.44 & 2.78 & 2.16 & 1.73 & 2.16 & 3.14 & 0.77 & 1.56 & 1.37 & 2.08 \\
FatesGS~\cite{han2025fatesgs}  & {1.32} & 2.85 & 2.71 & \best\textbf{0.80} & 1.44 & 2.08 & 1.11 & \best\textbf{1.19} & 1.33 & \underline{0.76} & 1.49 & \best\textbf{0.85} & \underline{0.47} & 1.05 & 1.06 & 1.37 \\
SolidGS~\cite{shen2024solidgs}  & 1.50 & \best\textbf{2.40} & 2.20 & 0.98 & \best\textbf{0.88} & \underline{1.36} & 0.79 & 1.29 & 1.33 & 1.06 & 1.72 & 1.07 & 0.60 & 0.99 & \best\textbf{0.89} & \underline{1.27} \\
\textbf{Ours(SfM-init)} & \underline{0.98} & 2.53 & 2.80 & 0.99 & 1.07 & 1.67 & 0.85 & 1.40 & 1.03 & 1.06 & 1.25 & 1.85 & 0.58 & 1.02 & 1.05 & 1.34 \\
\textbf{Ours(Epi-init)}  & \best\textbf{0.67} & \underline{2.58} & \best\textbf{1.72} & 0.88 & \underline{0.91} & \best\textbf{1.34} & \best\textbf{0.71} & 1.20 & \best\textbf{0.86} & \best\textbf{0.74} & \underline{1.06} & 1.00 & 0.54 & \underline{0.88} & \underline{0.94} & \best\textbf{1.07}  \\ \bottomrule
\end{tabular}}
\vspace{0pt}
\caption{\textbf{Quantitative results on DTU dataset (Sparse Little-Overlaps).} \textbf{Bold}-$1^{st}$, \underline{underline}-$2^{nd}$. Epi-init denotes initialization with epipolar depth priors.}
\vspace{-13pt}
\label{table:dtu_cd_little_overlap}
\end{table*}

\begin{table*}[t]
\centering
\renewcommand{\arraystretch}{1}
 \setlength{\tabcolsep}{5pt}
 \scalebox{0.65}{
\begin{tabular}{lcccccccccccccccc}
\toprule
Scan        & 24   & 37   & 40   & 55   & 63   & 65   & 69   & 83   & 97   & 105  & 106  & 110  & 114  & 118  & 122  & Mean \\ \midrule

Colmap~\cite{schonberger2016structure}      &0.90 &2.89 &1.63 &1.08 &2.18 &1.94 &1.61 &1.30 &2.34 &1.28 &1.10 &1.42 &0.76 &1.17 &1.14 &1.52 \\ \midrule

SparseNeus~\cite{long2022sparseneus}  &1.29	&2.27	&1.57	&0.88	&1.61	&1.86	&1.06	&1.27	&1.42	&1.07	&0.99	&0.87	&0.54	&1.15	&1.18	&1.27 \\					 
VolRecon~\cite{ren2023volrecon} &1.20	&2.59	&1.56	&1.08	&1.43	&1.92	&1.11	&1.48	&1.42	&1.05	&1.19	&1.38	&0.74	&1.23	&1.27 &1.38	  \\
UFORecon~\cite{youngju2024uforecon}        & 1.05 & 2.31 & 1.44 & 0.98 & 1.18 & 1.52 & 0.88 & 1.35 & 1.30 & 0.87 & 1.07 & 0.77 & 0.59 & 1.05 & 1.12 & 1.17 \\
NeuSurf~\cite{huang2024neusurf}   &\underline{0.78} &2.35 &1.55 &\best\textbf{0.75} &\best\textbf{1.04} &1.68 &\best\textbf{0.60} &1.14 &\underline{0.98} &\underline{0.70} &\best\textbf{0.74} &0.49 &\best\textbf{0.39} &\best\textbf{0.75} &\best\textbf{0.86} &0.99 \\ \midrule
2DGS~\cite{huang20242dgs}        &3.14	&3.79	&2.31	&1.51	&5.34	&2.22	&1.85	&2.79	&3.62	&1.63	&3.05	&4.11	&1.05	&3.35	&2.44	&2.81      \\
GS2Mesh~\cite{yaniv2024gs2mesh}        &1.34	&2.63	&1.45	&1.05	&2.90	&\underline{1.24}	&1.20	&1.26	&1.20	&0.79	&1.49	&0.70	&0.91	&1.08	&1.14	&1.36      \\
FatesGS~\cite{han2025fatesgs}         &\best\textbf{0.67}	&\underline{1.94}	&\best\textbf{1.17}	&\underline{0.77}	&1.28	&\best\textbf{1.23}	&\underline{0.63}	&\best\textbf{1.05}	&\underline{0.98}	&\best\textbf{0.69}	&\underline{0.75}	&\best\textbf{0.48}	&\underline{0.41}	&\underline{0.78}	&\underline{0.90}	&\best\textbf{0.92}     \\
Sparse2DGS~\cite{jiang2025sparse2dgs}    &1.05	&2.35	&1.38	&0.83	&1.37	&1.45	&0.84	&1.16	&1.43	&0.74	&0.85	&0.84	&0.57	&0.95	&1.01   &1.13
    \\ 
\textbf{Ours(Epi-init)}                       &0.84	&\best\textbf{1.51}	&\underline{1.18}	&0.92	&\underline{1.13}	&1.30	&0.70	&\underline{1.08}	&\best\textbf{0.95}	&0.75	&1.00	&\best\textbf{0.48}	&0.45	&0.92	&1.05   &\underline{0.95} \\ \bottomrule
\end{tabular}}
\vspace{0pt}
\caption{\textbf{Quantitative results on DTU dataset (Sparse Large-Overlaps).} \textbf{Bold}-$1^{st}$, \underline{underline}-$2^{nd}$. Epi-init denotes initialization with epipolar depth priors.}
\vspace{-10pt}
\label{table:dtu_cd_large_overlap}
\end{table*}

\begin{table*}[t] 
\centering
\begin{minipage}[t]{0.3\linewidth}
    \centering
    \vspace{-15pt} 
    \resizebox{\linewidth}{!}{
        \begin{tabular}{lc}
        \toprule
        Baseline & Chamfer Distance$\downarrow$ \\ 
        \midrule
        3\% & 1.240 \\
        7\% & 1.087 \\
        11\% & 1.078 \\
        15\% & 1.096 \\
        Adaptive Sel. & \textbf{1.069} \\ 
        \bottomrule
        \end{tabular}
    }
    \caption{\textbf{Reconstruction quality under different baselines} (\% of scene radius).}
    \label{table:baseline_ablation}
    \vspace{-6pt}
\end{minipage}
\hfill
\begin{minipage}[t]{0.62\linewidth}
    \centering
    \vspace{-15pt} 
    \resizebox{\linewidth}{!}{
        \begin{tabular}{c|ccc|ccc}
        \toprule
         & \multirow{2}{*}{$L_{stereo}$}
         & \multirow{2}{*}{$L_{co}$}
         & \multirow{2}{*}{$L_{depth}$}
         & \multicolumn{3}{c}{\textbf{Chamfer Distance$\downarrow$}} \\
        \cmidrule(lr){5-7}
        & &   &  & Little(SfM-init) & Little(Epi-init) & Large(Epi-init) \\
        \midrule
        (a) &   &      &    & 1.595 & 1.134  & 0.984 \\
        (b) & \ding{51} &    &  & 1.439 & 1.097  & 0.963 \\
        (c) & \ding{51} & \ding{51}  &  & 1.396 & 1.080  & 0.956 \\
        (d) & \ding{51} & \ding{51}  & \ding{51} & \textbf{1.341} & \textbf{1.069}  & \textbf{0.951} \\
        \bottomrule
        \end{tabular}
    }
    \caption{\textbf{Ablation of proposed components on DTU~\cite{aanaes2016dtu}.} Chamfer Distance evaluated across individual loss components and initialization strategies (Epi-init: epipolar depth).}
    \label{table:loss_ablation}
\end{minipage}
\vspace{-6pt}
\end{table*}

\begin{figure*}[t]
\centering
\includegraphics[scale=0.36]{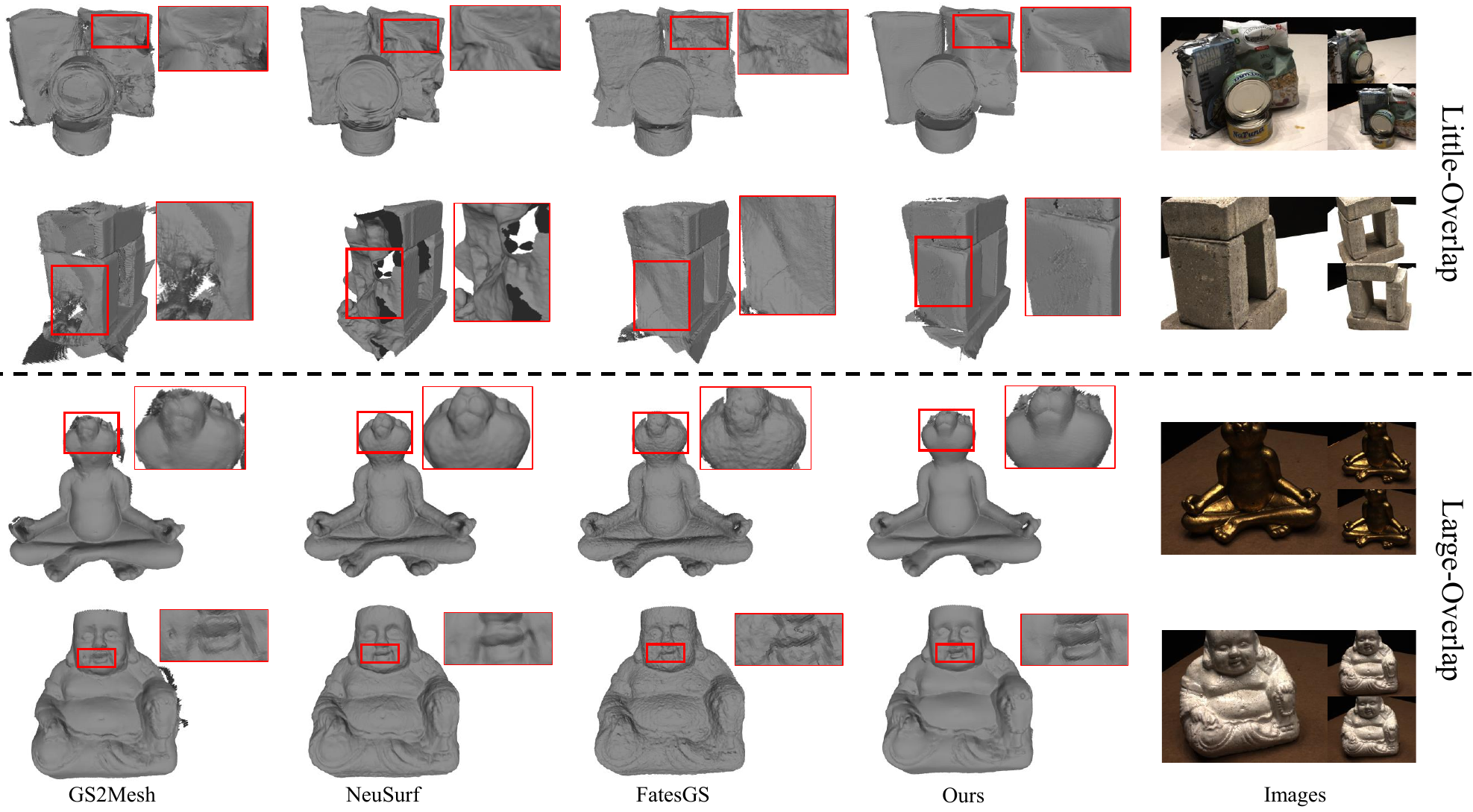}
\caption{\textbf{DTU surface reconstruction results under different sparse view settings.} Our method achieves more complete reconstruction with finer details.}
\label{fig:qualitative}
\vspace{-15pt}
\end{figure*}

\begin{figure}[!htbp]
\begin{center}
\includegraphics[scale=0.39]{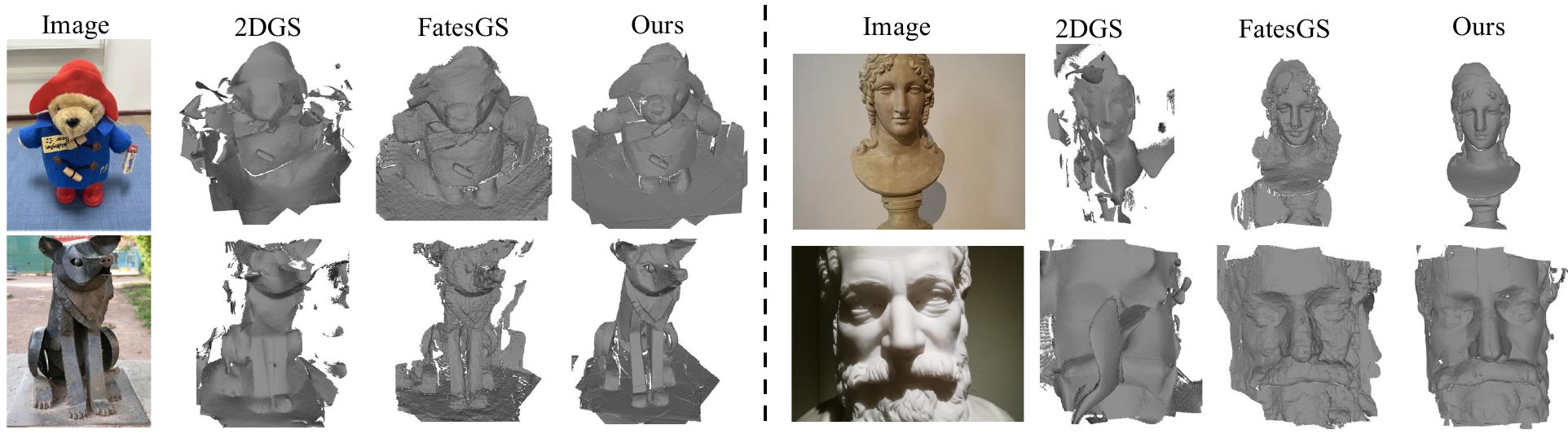}
\end{center}
\vspace{-10pt}
\caption{\textbf{BlendedMVS reconstruction results.} Our method demonstrates a more complete and detailed reconstruction.}
\label{fig:bmvs_results}
\vspace{-18pt}
\end{figure}

\subsection{Datasets}
\label{dataset}

We evaluate our method on 15 representative scenes from the DTU dataset~\cite{aanaes2016dtu} under both little-overlap (views 22, 25, 28)~\cite{alex2021pixelnerf} and large-overlap (views 23, 24, 33)~\cite{long2022sparseneus} settings. To evaluate generalization in real-world scenes, we perform qualitative comparisons on the BlendedMVS dataset~\cite{yao2020blendedmvs} using three randomly selected sparse views per scene. Following prior works, ground-truth camera poses are assumed to be known for all experiments

\subsection{Implementation Details}
\label{implementation}

Both 2DGS~\cite{huang20242dgs} and 3DGS~\cite{kerbl20233dgs} are warmed up for 7,000 iterations, followed by 1,000 iterations of fine-tuning before TSDF-based mesh extraction. Candidate baselines are sampled at 1\% intervals from 3\%--15\% of the scene radius for little-overlap, and 4\% intervals from 20\%--80\% for large-overlap settings. We employ Vis-MVSNet~\cite{zhang2020vismvsnet} for feature extraction $f_{\phi}$ and DEFOM-Stereo~\cite{jiang2025defomstereo} as the stereo network $f_{\psi}$. During inference, occluded regions ($(M_{occ}^{tr-st})^\mathrm{c}$) are completed using epipolar depth priors from initialization.
\subsection{Experiment Results}
\label{result}

We evaluate surface reconstruction on the DTU dataset using Chamfer Distance (CD) under both little-overlap (Tab.~\ref{table:dtu_cd_little_overlap}) and large-overlap (Tab.~\ref{table:dtu_cd_large_overlap}) settings against SfM-, NeRF-, and 3DGS-based baselines. With epipolar depth initialization, our method achieves SOTA performance in the little-overlap setting, outperforming the previous best method SolidGS~\cite{shen2024solidgs} by 15\%, while maintaining competitive performance under the large-overlap setting. Even with standard SfM initialization, our model remains robust and competitive, demonstrating that our gains do not solely rely on a specific prior. Qualitatively, our framework preserves finer geometric details and global consistency on DTU (Fig.~\ref{fig:qualitative}) and generalizes effectively to challenging real-world scenes in BlendedMVS~\cite{yao2020blendedmvs} (Fig.~\ref{fig:bmvs_results}).

\subsection{Ablation Studies}
\label{ablation}
\paragraph{Adaptive Baseline Selection.}  
To evaluate the effectiveness of the proposed adaptive baseline selection, we compare our method against fixed settings where the baseline is manually set to 3\%, 7\%, 11\%, and 15\% of the scene radius across all views.
As shown in Tab.\ref{table:baseline_ablation}, our adaptive baseline selection automatically determines optimal baseline values for each view without any manual tuning, achieving lower Chamfer Distance compared to fixed baselines.
Importantly, the primary contribution is not merely performance improvement, but the elimination of manual hyperparameter search. Manual baseline tuning requires running all candidate configurations, which takes approximately one hour per scene, whereas our adaptive strategy determines suitable baselines within one minute.

\paragraph{Proposed Components in Fine-Tuning.} 

To evaluate the contribution of each component, we isolate individual modules on the DTU (Tab.~\ref{table:loss_ablation}). While Epi-init provides a strong prior with steady improvements (5.7\% gain), ablations with SfM-init demonstrate a substantial 15.9\% gain, clearly validating the effectiveness and novelty of our proposed framework under weak initialization. Specifically, backpropagating depth-refining gradients from the stereo matching network optimizes 3DGS for stereo matching (row \textbf{b}). Furthermore, the 2D/3D co-regularization leverages their complementary strengths to stabilize optimization (row \textbf{c}), and enforcing depth consistency between the stereo model and 2DGS achieves the best reconstruction accuracy across both overlap settings (row \textbf{d}).


\section{Conclusions}

In this paper, we present Sparse-GS2Mesh, a framework integrating Gaussian splatting and stereo matching for sparse-view surface reconstruction. We stabilize optimization by leveraging epipolar depth priors for initialization and automatically determining baselines via adaptive baseline selection. Furthermore, fine-tuning with a stereo matching network and 2D/3D co-regularization specializes 3DGS to generate stereo-consistent renderings that provide robust geometric cues. Rather than a naive combination, our components explicitly complement each other to resolve incomplete geometry. Consequently, our method demonstrates strong robustness across diverse sparse-view regimes and achieves state-of-the-art reconstruction quality on DTU and BlendedMVS datasets.

\newpage
\section*{Acknowledgements}
This work was supported by Institute of Information \& communications Technology Planning \& Evaluation (IITP) grant funded by the Korea government
(MSIT) (No. RS-2026-25522885, Development of a World Foundation Model for
Training and Deployment of Physical AI Systems).

\bibliographystyle{plain}
\bibliography{main}

\newpage
\appendix

\section{Computational Report}
Our framework employs a foundation stereo~\cite{jiang2025defomstereo} whose GPU memory usage is highly sensitive to input resolution.
We evaluate this trade-off on DTU under different input image downsampling ratios, reporting training time, GPU memory, and reconstruction accuracy.
As shown in Tab.\ref{table:computational_report}, 50\% resolution reduces memory usage to 21 GB with only a 0.07 increase in Chamfer Distance compared to full resolution, substantially outperforming the previous state of the art (1.27)~\cite{shen2024solidgs}.
This enables flexible resolution adjustment based on available GPU memory with minimal quality loss.
The total training time is at most 37 minutes, including 20 minutes of warm-up, which is slightly longer than GS-based baselines~\cite{han2025fatesgs, jiang2025sparse2dgs} but significantly faster than NeRF-based methods~\cite{huang2024neusurf, youngju2024uforecon}, while achieving better performance.
\begin{table}[htbp]
\centering
\renewcommand{\arraystretch}{1}
 \setlength{\tabcolsep}{12pt}
\scalebox{0.8}{
\begin{tabular}{cccc}
\toprule
\textbf{Downsample}    & \textbf{GPU Memory} & \textbf{Training Time} & \textbf{CD$\downarrow$} \\ \midrule
50\%    &     21GB    &   31 mins  &  1.140   \\
66\%  &     34GB    &   33 mins  &   1.117   \\
100\%      &     73GB    &   37 mins  & \textbf{1.069}  \\
\bottomrule
\end{tabular}}
\vspace{5pt}
\caption{Computational report across downsampling ratios.}
\vspace{-5pt}
\label{table:computational_report}
\end{table}

\section{Different Initialization Strategies.}
\begin{figure}[H]
\begin{center}
\includegraphics[scale=0.54]{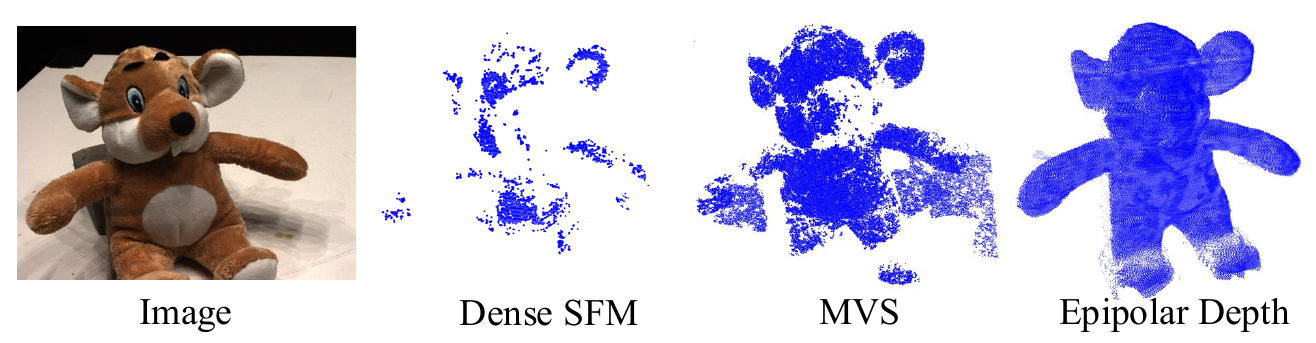}
\end{center}
\vspace{-15pt}
\caption{\textbf{Generated Initial Pointcloud.} The point cloud generated from epipolar depth is the most dense and preserves the finest geometric details}
\label{fig:initialization}
\vspace{-10pt}
\end{figure}

We conduct an additional experiment to analyze the difference in initialization strategies for the 3DGS~\cite{kerbl20233dgs} and 2DGS~\cite{huang20242dgs} modules that affect the final reconstruction quality.
We compare three sources of point clouds: (1) SfM~\cite{schonberger2016structure} point clouds obtained from only three views, (2) point clouds generated by MVSNet~\cite{cao2023mvsformer} under sparse-view settings, and (3) point clouds constructed from our chosen epipolar depth~\cite{yulong2025nexusgs}.
As shown in Fig.\ref{fig:initialization}, the point cloud produced from epipolar depth preserves the finest geometric details among all comparisons, demonstrating its effectiveness as an initialization strategy.

\begin{figure}[H]
\begin{center}
\includegraphics[scale=0.51]{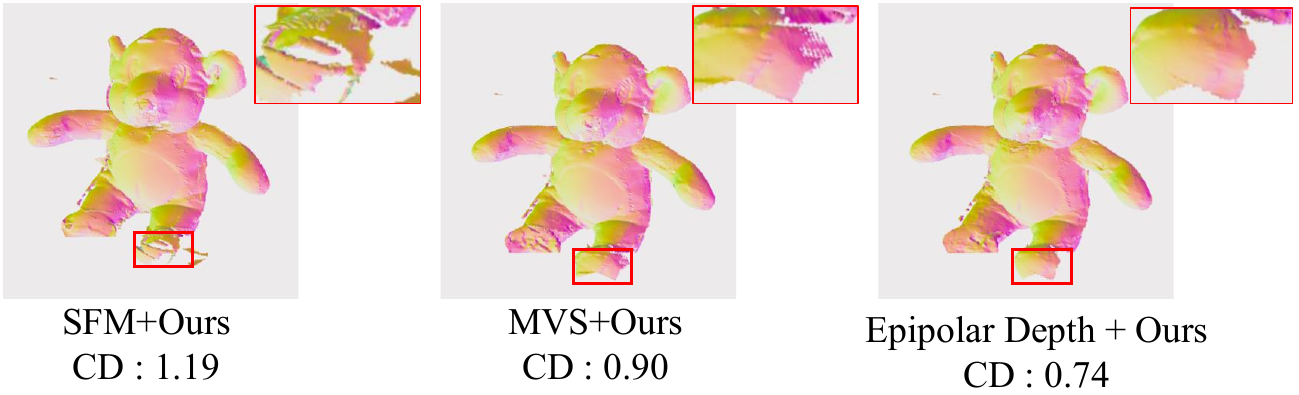}
\end{center}
\vspace{-1pt}
\caption{\textbf{Visual comparison of the final reconstructed meshes under different initialization strategies.} }
\label{fig:initialization_mesh}
\vspace{-1pt}
\end{figure}

This experiment is conducted under the little-overlap setting, and all configurations except for the initialization method follow the same setup aforementioned.
The corresponding mesh reconstruction quality for each initialization method is reported in Tab.\ref{table:intialization_comparsion}.
Initializing with epipolar depth yields the best performance overall.
However, the point clouds generated by MVSNet also produce results that surpass the previous state-of-the-art method, SolidGS (1.27), indicating that our method remains robust regardless of the specific initialization strategy.
Moreover, as shown in Fig.\ref{fig:initialization_mesh}, although the epipolar-depth initialization produces the finest geometric details, the MVSNet-based initialization also reconstructs sufficiently reasonable geometry, further demonstrating the robustness of our approach.

\begin{table}[H]
\centering
\renewcommand{\arraystretch}{1}
 \setlength{\tabcolsep}{12pt}
\scalebox{1.0}{
\begin{tabular}{lccc}
\toprule
Init. Strategy & Dense SFM~\cite{schonberger2016structure} & MVSNet~\cite{cao2023mvsformer} & Epipolar Depth~\cite{yulong2025nexusgs}  \\
\midrule
CD$\downarrow$   
       &      1.341     
       &    1.123     
      &     \textbf{1.069}    \\

\bottomrule
\end{tabular}}
\vspace{5pt}
\caption{\textbf{Reconstruction quality of our method under different initialization strategies.}}
\vspace{-1pt}
\label{table:intialization_comparsion}
\end{table}

\section{Stereo View Image Rendering.} 
\begin{figure}[htbp]
\begin{center}
\includegraphics[scale=0.54]{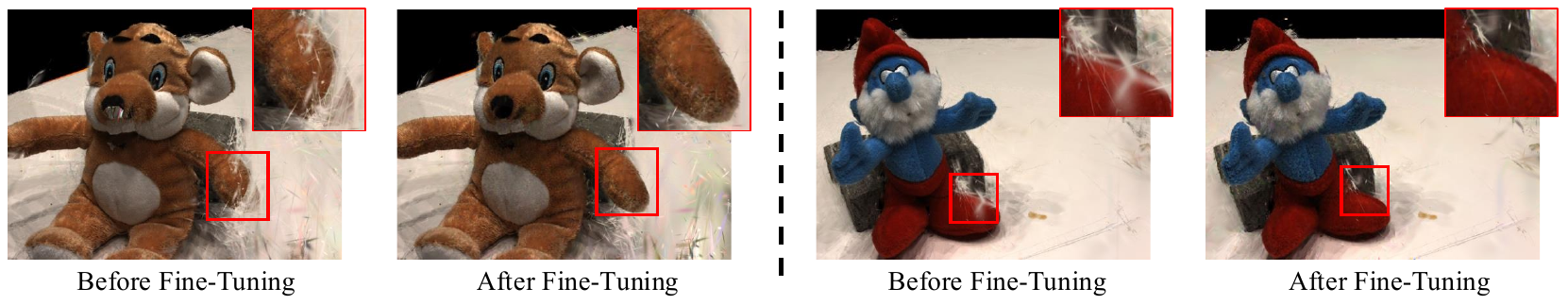}
\end{center}
\vspace{-8pt}
\caption{\textbf{Effect of fine-tuning on rendering quality.} After the fine-tuning, artifacts are removed, resulting in more accurate stereo renderings.}
\label{fig:stereo_view_rendering}
\vspace{-10pt}
\end{figure}

We compare the quality of stereo view rendering before and after fine-tuning under the stereo view determined in Sec \ref{adaptive baseline determination}.
Since the stereo views do not have ground truth images, we train the 3DGS under the dense-view setting and use the rendered results as pseudo ground truth for evaluation.
We evaluate PSNR and LPIPS, measuring the performance on the occlusion mask $M_{occ}^{tr-st}$, which is affected by the gradients of our method.
As shown in Tab.\ref{table:stereoviewrenderingquality}, both metrics improve after fine-tuning, demonstrating the effectiveness of our optimization.
Moreover, as illustrated in Fig.\ref{fig:stereo_view_rendering}, visual artifacts observed in the stereo views before fine-tuning are removed after optimization, resulting in more consistent renderings.

\begin{table}[!htbp]
\centering
\renewcommand{\arraystretch}{1}
 \setlength{\tabcolsep}{12pt}
\scalebox{0.92}{
\begin{tabular}{lcc}
\toprule
Fine-Tuning    & PSNR$\uparrow$ & LPIPS$\downarrow$ \\ \midrule
Before       &     26.27         & 0.058  \\
After        &     \textbf{26.57}     &    \textbf{0.054}   \\
\bottomrule
\end{tabular}}
\vspace{5pt}
\caption{\textbf{Rendering quality comparison before and after proposed fine-tuning on stereo view.} The image quality improves after fine-tuning, showing more accurate stereo renderings.}
\label{table:stereoviewrenderingquality}
\end{table}

\end{document}